\documentclass[conference]{IEEEtran}
\IEEEoverridecommandlockouts
\usepackage{cite}
\usepackage{amsmath,amssymb,amsfonts}

\usepackage{algorithm}
\usepackage{algpseudocode}

\usepackage{graphicx}
\usepackage{multirow}
\usepackage{multicol}
\usepackage{url}
\usepackage{booktabs}
\usepackage{listings}

\usepackage{hyperref}

\def\BibTeX{{\rm B\kern-.05em{\sc i\kern-.025em b}\kern-.08em
    T\kern-.1667em\lower.7ex\hbox{E}\kern-.125emX}}
\begin{document}



\title{xLSTMAD: A Powerful xLSTM-based Method \\ for Anomaly Detection\thanks{Paper accepted at ICDM2025}}
%
%
%

%
%

\author{\IEEEauthorblockN{Kamil Faber\IEEEauthorrefmark{1}, Marcin Pietron\IEEEauthorrefmark{1}, Dominik Zurek\IEEEauthorrefmark{1}, Roberto Corizzo\IEEEauthorrefmark{2}}
\IEEEauthorblockA{\IEEEauthorrefmark{1}AGH University of Krakow, Poland \\
Email: kfaber@agh.edu.pl, pietron@agh.edu.pl, dzurek@agh.edu.pl}
\IEEEauthorblockA{\IEEEauthorrefmark{2}American University, Washington DC, USA\\
Email: rcorizzo@american.edu}}
 

\maketitle

\begin{abstract}
The recently proposed xLSTM is a powerful model that leverages expressive multiplicative gating and residual connections, providing the temporal capacity needed for long-horizon forecasting and representation learning. This architecture has demonstrated success in time series forecasting, lossless compression, and even large-scale language modeling tasks, where its linear memory footprint and fast inference make it a viable alternative to Transformers. Despite its growing popularity, no prior work has explored xLSTM for anomaly detection. In this work, we fill this gap by proposing \texttt{xLSTMAD}, the first anomaly detection method that integrates a full encoder-decoder xLSTM architecture, purpose-built for multivariate time series data. Our encoder processes input sequences to capture historical context, while the decoder is devised in two separate variants of the method. In the forecasting approach, the decoder iteratively generates forecasted future values \texttt{xLSTMAD-F}, while the reconstruction approach reconstructs the input time series from its encoded counterpart \texttt{xLSTMAD-R}. We investigate the performance of two loss functions: Mean Squared Error (MSE), and Soft Dynamic Time Warping (SoftDTW) to consider local reconstruction fidelity and global sequence alignment, respectively. We evaluate our method on the comprehensive TSB-AD-M benchmark, which spans 17 real-world datasets, using state-of-the-art challenging metrics such as VUS-PR. In our results, xLSTM showcases state-of-the-art accuracy, outperforming 23 popular anomaly detection baselines. 
Our paper is the first work revealing the powerful modeling capabilities of xLSTM for anomaly detection, paving the way for exciting new developments on this subject. Our code is available at: \url{https://github.com/Nyderx/xlstmad}.

\end{abstract}


\section{Introduction}

Anomaly detection in multivariate time series is a vital task across domains such as industrial monitoring, cybersecurity, human activity recognition, and space system diagnostics. While statistical techniques and classical machine learning models remain strong baselines, deep learning methods have emerged as powerful alternatives due to their ability to model complex temporal dependencies and high-dimensional patterns. Recent benchmarks like TSB-AD \cite{liu2024elephant} reveal that even simple neural architectures (e.g., CNNs \cite{munir2018deepant}, LSTMs \cite{ji2021novel}) can outperform more complex Transformer variants \cite{tuli2022tranad,xuanomaly} in anomaly detection tasks—particularly in multivariate settings where expressive temporal modeling is key.

Most state-of-the-art deep learning approaches fall into two broad categories: reconstruction-based models (e.g., autoencoders, VAEs) that detect anomalies through high reconstruction error, and prediction-based models (e.g., LSTMs) that forecast future behavior and identify deviations. However, these models often struggle to balance long-term memory, fine-grained temporal resolution, and computational efficiency \cite{liu2024elephant}.

The recently proposed xLSTM architecture offers a promising research direction in addressing all these gaps  \cite{beck2025xlstm}. As a generalization of LSTM, it combines depth-wise convolutions, residual connections, and multi-scale gating mechanisms in a stackable, efficient structure. xLSTM has demonstrated success in time series forecasting \cite{auer2025tirexzeroshotforecastinglong}, lossless compression \cite{ma2025multi}, and even large-scale language modeling \cite{beck2025xlstm7b}, where its linear memory footprint and fast inference make it a viable Transformer alternative. Despite its growing popularity, no prior work has explored xLSTM for anomaly detection.

In this work, we fill this gap by proposing \texttt{xLSTMAD}, the first encoder-decoder xLSTM model for multivariate time series anomaly detection. Our architecture is purpose-built: an encoder processes input sequences to capture historical context, while a decoder is devised in two separate variants. In the forecasting approach, the decoder iteratively generates forecasted future values \texttt{xLSTMAD-F}, while the reconstruction approach reconstructs the input time series from its encoded counterpart \texttt{xLSTMAD-R}. 
We investigate the performance of two loss functions: Mean Squared Error (MSE), and Soft Dynamic Time Warping (SoftDTW), with the intention to consider local reconstruction fidelity and global sequence alignment, respectively. 

To rigorously evaluate our approach, we leverage the TSB-AD-M benchmark \cite{liu2024elephant}, a comprehensive suite of 17 real-world datasets ranging from industrial control systems (e.g., SWaT, SMD) to spacecraft telemetry (e.g., SMAP, MSL) and human activity monitoring (e.g., OPP), designed to create extremely challenging conditions that are useful to assess the effectiveness of diverse models in diverse conditions. 
Our results show that the xLSTM encoder-decoder architecture achieves superior performance compared to classical and modern baselines. 


The contributions of this work can be summarized as:
\begin{itemize}
\item{We propose \texttt{xLSTMAD}, the first encoder-decoder architecture based entirely on xLSTM blocks for multivariate time series anomaly detection, leveraging its efficient recurrence, convolutional depth, and long-range memory.}
\item{The proposed \texttt{xLSTMAD} method integrates residual connections, depthwise convolutions, and a hybrid stacking of mLSTM and sLSTM blocks. The model is designed to balance expressiveness and stability, with selective placement of gating and feedforward components.}
\item{We propose and evaluate two variants of \texttt{xLSTMAD}:  \textit{forecasting-based} anomaly scoring \texttt{xLSTMAD-F}, and \textit{reconstruction-based} \texttt{xLSTMAD-R}, allowing flexibility in how the decoder’s outputs are used to identify abnormal behavior depending on the domain or data characteristics.}
\item{We investigate the performance of two loss functions:  \textit{Mean Squared Error (MSE)} and \textit{Soft Dynamic Time Warping (SoftDTW)}, which capture pointwise deviations and temporal distortions, crucial for detecting subtle or delayed anomalies, respectively.}
\item{We evaluate our method on 17 real-world datasets from the TSB-AD-M benchmark, the most challenging time series anomaly detection benchmark covering diverse domains such as industrial control, spacecraft telemetry, and human activity, and resorting to comprehensive metrics such as VUS-PR, in addition to more traditional metrics.} 
\end{itemize}


\section{Background}

\subsection{Time series anomaly detection}

Anomaly detection in time series encompasses a range of classical and modern methods. Traditional statistical approaches assume an underlying generative model of normal behavior and use statistical tests or thresholds to flag outliers. For example, one may fit an ARIMA or seasonal model and then apply hypothesis tests on residuals~\cite{aggarwal2017introduction}. 
%
Clustering and density-based methods form another classical category. These unsupervised approaches group similar time-series segments (e.g., via k-means or hierarchical clustering) and define a normal ,,profile'' of data. Points that lie far from cluster centroids or form very small clusters are labeled anomalous~\cite{aggarwal2017introduction,breunig2000lof}. Likewise, distance-based techniques (e.g., k-NN, LOF) and density-based methods (e.g., DBSCAN) treat points in sparse regions of feature space as outliers
~\cite{aggarwal2017introduction,li2020copod}.

In the deep learning era, reconstruction-based models have become popular for time-series anomaly detection. Autoencoders (AEs) are neural networks trained to reconstruct input sequences, and anomalies are identified by high reconstruction error. Sakurada and Yairi~\cite{sakurada2014anomaly} showed that even simple AEs can outperform linear PCA for detecting anomalies in multivariate time series. Modern variants (e.g., denoising AEs, convolutional AEs) follow the same principle. For instance, the USAD model~\cite{audibert2020usad} uses two adversarially-trained autoencoders: by amplifying reconstruction error, it more clearly separates anomalous from normal patterns in multivariate data. In general, these methods learn a low-dimensional embedding of the time series and score anomalies by how poorly the model can reconstruct a given sequence.

Generative deep models explicitly learn the probability distribution of normal time-series data. Variational autoencoders (VAEs) and GANs are common in this class. For example, OmniAnomaly~\cite{su2019robust} uses a recurrent VAE with planar normalizing flows to model multivariate time series: it learns a compact latent representation of normal behavior and flags anomalies when the reconstruction probability is low. GAN-based schemes similarly train a generative model on normal sequences and use the generator or discriminator to score anomalies (e.g., a low discriminator confidence or high reconstruction loss indicates an anomaly).


Recently, transformer and attention-based architectures have been applied to anomaly detection. Transformers process the entire sequence via self-attention, capturing long-range dependencies without explicit recurrence. The Anomaly Transformer~\cite{xuanomaly} uses self-attention to learn pairwise associations between time points: it models both ``prior'' (local Gaussian-weighted) and ``series'' (learned self-attention) associations, and anomalies are detected when these expected associations are disrupted.  Related models like TranAD~\cite{tuli2022tranad} demonstrate that transformer-based models can differentiate anomalies by their deviant attention patterns and reconstruction discrepancies.





\subsection{xLSTM}
xLSTM is rapidly gaining traction in different tasks. The most natural application as a direct successor of the LSTM architecture is in time series prediction tasks. 
The work in \cite{fan2024advanced} proposed an advanced stock price prediction model based on xLSTM, with the aim of enhancing predictive accuracy in both short and long term periods. Experimental results with multiple stocks showed that xLSTM consistently outperforms LSTM across all stocks and time horizons, and this gap widens as the prediction period extends.

The success of xLSTM in time series prediction led researchers to make efforts towards its adaptation towards other tasks, data types, and domains.
%
An example is lossless compression based on neural networks. The work in \cite{ma2025multi} propose a novel lossless compressor with two compression stages:  parallel expansion mapping, which maps redundant pieces in multi-source data into unused alphabet values, and compression through an xLSTM model with a Deep Spatial Gating Module (DSGM). 
%
Authors in \cite{alkin2024vision}  introduce Vision-LSTM (ViL), an adaptation of xLSTM building blocks to computer vision. The architecture provides a stack of xLSTM blocks in which odd blocks process the
sequence of patch tokens from top to bottom while even blocks go from bottom to top. The proposed model achieved strong classification performance in transfer learning and segmentation tasks, showing promise for further deployment of xLSTM in computer vision architectures.
%
Similarly, LLMs built on the xLSTM architecture have emerged as a viable alternative to Transformers, due to a linear relationship between sequence length and compute requirements, as well as the constant memory usage. 
The authors in \cite{beck2025xlstm7b} introduced xLSTM 7B, a 7-billion-parameter LLM with targeted optimizations for fast and efficient inference. 
The model was able to achieve performance on downstream tasks similar to popular models such as  Llama- and Mamba-based LLMs with a significantly faster inference speed \cite{beck2025xlstm7b}. 

Despite the recent success of xLSTM in many contexts, to the best of our knowledge, no study thus far has explored its capabilities for time series anomaly detection. The goal of our paper is to address this gap.

\subsection{TSB-AD anomaly detection benchmark}

In the original TSB-AD study \cite{liu2024elephant}, a comprehensive collection of 40 time-series anomaly detection algorithms is leveraged to compare statistical, neural network-based, and the latest foundation model-based methods. 
Extensive experiments conducted on the TSB-AD benchmark led to relevant insights.
First, statistical methods generally exhibit strong and consistent performance. In contrast, neural network-based models often fall short of their reputed superiority, though they still perform competitively in detecting point anomalies and handling multivariate time series.
Moreover, simple architectures like CNNs and LSTMs tend to outperform more sophisticated designs such as advanced Transformer variants. 
%
%
%

%
Even though CNN \cite{munir2018deepant} and OmniAnomaly \cite{su2019robust} display the strongest performance on multivariate datasets, the analysis provided by the TSB-AD benchmark highlights the need for more expressive models in complex multivariate settings.

\section{Method}
\begin{figure}[h!]
\centering
\includegraphics[width=\columnwidth]{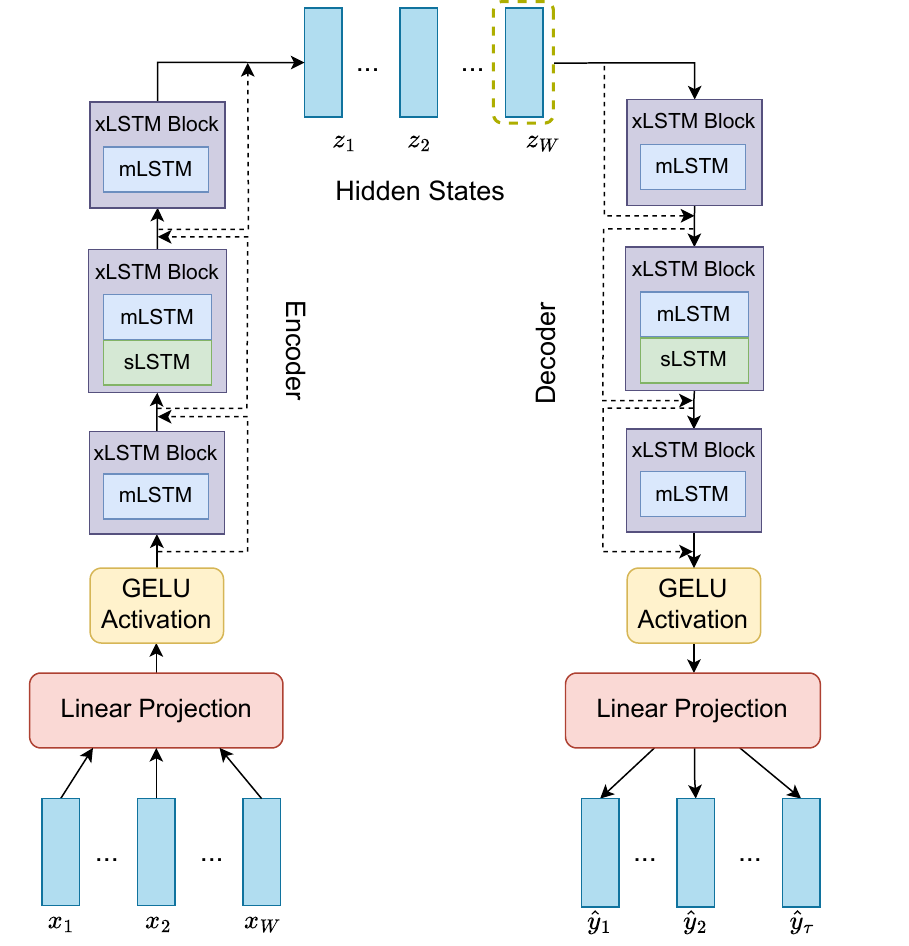}
\caption{Proposed \texttt{xLSTMAD} Encoder-Decoder model architecture. Input data is organized as multivariate time series $x_1, x_2, \dots, x_W$, where $W$ is the window size. The data fed to the Encoder is processed into hidden states $z_1, z_2, \dots, z_W$ through linear projection, GELU activation, and multiple residually stacked xLSTM blocks consisting of mLSTM and optional sLSTM layers. The Decoder generates reconstructions \texttt{xLSTMAD-R}or forecasts \texttt{xLSTMAD-F}, which are then leveraged in the computation of anomaly scores. Dashed lines denote residual connections between stacked blocks. 
}
\label{fig:model}
\end{figure}

In this section, we describe our proposed \texttt{xLSTMAD} method. The model architecture is shown in Figure \ref{fig:model}. In the following, we first formally introduce xLSTM, the foundational layer used in our architecture. Second, we showcase the model architecture in terms of its constituent blocks. Third, we devise the two variants of our method: forecasting \texttt{xLSTMAD-F} and reconstruction-based \texttt{xLSTMAD-R}. Finally, we explain the two losses adopted to train the model architecture.

\subsection{xLSTM}
The \textit{Extended Long Short-Term Memory (xLSTM)} \cite{beck2025xlstm} model architecture presents a significant advancement over the classical LSTM by addressing three major limitations: \textit{i)} inability to revise past storage decisions, \textit{ii)} limited memory capacity due to scalar-valued cell states, and \textit{iii)} lack of parallelizability stemming from sequential hidden-to-hidden state dependencies.
Specifically, a standard LSTM updates its memory cell as:
\begin{equation}
    c_t = f_t c_{t-1} + i_t z_t,
\end{equation}
where $f_t$, $i_t$ are sigmoid gates and $z_t$ is the candidate memory input. In contrast, xLSTM introduces \textit{exponential gating}, where gates can take the form:
\begin{equation}
    i_t = \exp(\tilde{i}_t), \quad f_t = \exp(\tilde{f}_t),
\end{equation}
allowing for a sharper and more dynamic weighting of memory contributions. A normalization mechanism via a \textit{normalizer state} $n_t$ ensures numerical stability:
\begin{equation}
    n_t = f_t n_{t-1} + i_t, \quad \tilde{h}_t = \frac{c_t}{n_t}.
\end{equation}
This behavior enables effective revision of prior information. We argue that this property is beneficial for detecting anomalous patterns that require continual memory updates as new evidence arrives.

\textit{Memory Mixing:} xLSTM architecture introduces a new \textbf{sLSTM} cell with a scalar memory, a scalar update, and memory mixing. As exponential activation functions can lead to large values that cause
overflows, the sLSTM cell includes an additional state $m_t$ leveraged to stabilize the gates. Moreover, sLSTM leveraged a new memory mixing approach, in which a hidden state vector is connected to the memory cell and input, forgetting, and output gates. Moreover, while sLSTM can have multiple heads, the memory mixing is adopted only within each head, not across different heads. 



\vspace{5pt}

\textit{Matrix-Valued Memory}: To overcome the bottleneck of scalar memory cells, xLSTM introduces the \textbf{matrix LSTM (mLSTM)} with a memory update rule:
\begin{equation}
    \boldsymbol{C}_t = f_t \boldsymbol{C}_{t-1} + i_t \boldsymbol{v}_t \boldsymbol{k}_t^\top,
\end{equation}
where $\boldsymbol{v}_t$ and $\boldsymbol{k}_t$ are value and key vectors, respectively. This outer-product update—similar in spirit to associative memories—enables richer storage and retrieval capabilities, which are crucial for handling high-dimensional or temporally dispersed anomalies in time series.

\vspace{5pt}

\textit{Parallelizable Design}: Unlike traditional LSTMs, the mLSTM variant included in xLSTM removes hidden-to-hidden connections, making it fully parallelizable. This addresses the LSTM’s sequential bottleneck and improves scalability, allowing deployment in large-scale, real-time anomaly detection systems.

These upgrades are particularly relevant for anomaly detection in time series data, where models must: \textit{i)} retain and revise long-range dependencies (enabled by exponential gating), \textit{ii)} capture complex, multi-dimensional patterns (through matrix-valued memory), and \textit{iii)} scale efficiently to long sequences and high-frequency streams (via parallelizable updates).
xLSTM thus provides a more expressive, adaptive, and efficient sequence modeling backbone for anomaly detection tasks compared to conventional LSTMs.
However, thus far, no research on xLSTM has been carried out in the realm of anomaly detection. The main goal of our study is to investigate the effectiveness of xLSTM models for time series anomaly detection tasks.

\subsection{Proposed model architecture}
Given an input time series $\mathbf{X} \in \mathbb{R}^{B \times W\times F}$, where $B$ is the batch size, $W$ is the context (window) length, and $F$ is the input features dimension, the model processes the sequence through three stages described below.


\vspace{5pt}

\subsubsection*{1. Input Projection}

The input is projected into a $D$-dimensional embedding space:
\begin{equation}
\mathbf{H}_0 = \phi(\mathbf{X} \mathbf{W}_p + \mathbf{b}_p) \in \mathbb{R}^{B \times W \times D}
\end{equation}
where $\mathbf{W}_p \in \mathbb{R}^{F \times D}$ and $\phi$ is a non-linear activation (e.g., GELU).

\vspace{5pt}
\subsubsection*{2. Encoder: Residual xLSTM Blocks}
xLSTM encoder $\texttt{xLSTM}_{enc}$) is built of 3 xLSTM blocks, each of which consists of an mLSTM layer with an optional sLSTM layer.
For simplicity, in the following, we describe how a single xLSTM block of $\texttt{xLSTM}_{enc}$ is defined.

Let the encoder contain $L$ stacked blocks, indexed by $\ell = 1, \dots, L$. 
For each layer $\ell$, the output of an mLSTM layer is defined as:
\begin{equation}
\mathbf{H}_\ell^{(m)} = \text{mLSTMCell}_\ell(\text{Conv1D}_8(\mathbf{H}_{\ell-1}))
\end{equation}
where $\text{Conv1D}_8$ corresponds to one dimensional convolutional layer with kelner size = 8.

If sLSTM layer is included in the block $\ell$, we define its output as:
\begin{equation}
\mathbf{H}_\ell^{(s)} = \text{sLSTMCell}_\ell(\text{Conv1D}_4(\mathbf{H}_\ell^{(m)}))
\end{equation}
\begin{equation}
\mathbf{H}_\ell^{(f)} = \text{FFN}_\ell(\mathbf{H}_\ell^{(s)}) \\
\end{equation}



The output of a single xLSTM block leverages the output of the sLSTM layer and the residual connection:
\begin{equation}
\mathbf{H}_\ell = \mathbf{H}_{\ell-1} + \mathbf{H}_\ell^{(f)}
\end{equation}



\vspace{5pt}
\subsubsection*{3. Decoder: Autoregressive Forecasting}
Following the encoding stage, the decoding stage generates forecasts for subsequent time steps \texttt{xLSTMAD-F} or multi-dimensional reconstructions \texttt{xLSTMAD-R}, depending on the preferred variant.

The decoding stage includes a decoder \texttt{$xLSTM_{dec}$} consisting of multiple xLSTM blocks, similar to \texttt{$xLSTM_{enc}$} defined above.


First, the decoder state is initialized from the last encoder timestep:
\begin{equation}
    \mathbf{h}_0 = \mathbf{H}_{L}[:, -1, :] \in \mathbb{R}^{B \times D}
\end{equation}

Second, predictions are generated for $t=1, \dots, \tau$:
\begin{equation}
\mathbf{h}_t = \texttt{xLSTM}_{\text{dec}}(\mathbf{h}_{t-1})
\end{equation}
where $\tau$ is equal to the forecasting horizon in \texttt{xLSTMAD-F} and to the original sequence size in reconstruction variant xLSTADM-R.

Finally, a GELU activation function $\phi$ is applied to the decoded hidden state, and the result is projected into the $F$-dimensional feature space:
\begin{equation}
\hat{\mathbf{y}}_t = \phi(\mathbf{z}_t \mathbf{W}_o + \mathbf{b}_o )\in \mathbb{R}^{B \times F}
\end{equation}

The final output sequence can be formalized as:
\begin{equation}
\hat{\mathbf{Y}} = [\hat{\mathbf{y}}_1, \hat{\mathbf{y}}_2, \dots, \hat{\mathbf{y}}_\tau] \in \mathbb{R}^{B \times \tau \times F}
\end{equation}


\subsection{Anomaly detection strategies}
%
%
Let us recall that for a given time step $t$, the model takes as input the historical window of size $W$:
\begin{equation}
\mathbf{X}_{t}^{(W)} = [\mathbf{x}_{t-W+1}, \ldots, \mathbf{x}_t] \in \mathbb{R}^{W \times F}
\end{equation}
We propose two anomaly detection approaches: Forecasting \texttt{xLSTMAD-F} and Reconstruction \texttt{xLSTMAD-R}. 

\vspace{5pt}

\subsubsection{Forecasting \texttt{xLSTMAD-F}}



For a given time step $t$, the model takes as input the historical window $\mathbf{X}_{t}^{(W)}$
and generates forecasts for the next $p$ steps, where $p$ defines the forecasting horizon:
\begin{equation}
\widehat{\mathbf{X}}_{t+1:t+p} = f_\theta(\mathbf{X}_{t}^{(W)}) \in \mathbb{R}^{p \times D}
\end{equation}

The ground truth for the prediction is:
\begin{equation}
\mathbf{X}_{t+1:t+p} = [\mathbf{x}_{t+1}, \mathbf{x}_{t+2}, \ldots, \mathbf{x}_{t+p}]
\end{equation}
In principle, using multiple prediction time steps should allow the model to leverage the smoothness of multi-step-ahead forecasts, with a potential improved anomaly detection robustness. 
%
However, this setup assumes that anomalies will be detected with some delay, since, after prediction, the computation of the anomaly score requires the next $p$ time steps. Therefore, the choice of forecasting horizon $p$ may depend on a specific domain and requirements.
%

The pointwise prediction error can be quantified using the Mean Squared Error (MSE) across the prediction horizon and all dimensions:
\begin{equation}
\mathcal{L}_{\text{pred}}(t) = \frac{1}{p \cdot D} \sum_{i=1}^{p} \sum_{j=1}^{D} \left(x_{t+i, j} - \widehat{x}_{t+i, j} \right)^2
\label{eq:L_pred}
\end{equation}

This prediction error is used directly as the raw anomaly score at time $t + p$, indicating how well the model's forecast matches the actual future behavior, thus enabling a threshold-free evaluation setup.


The forecasting-based approach enables the model to directly learn temporal dynamics and causal dependencies, making it effective for detecting subtle deviations from expected future behavior. Such deviations often signal contextual or collective anomalies in the sequence.

\subsubsection{Reconstruction \texttt{xLSTMAD-R}}
In the reconstruction-based paradigm, we train a model to learn a compressed representation of past observations and reconstruct them as accurately as possible.


A reconstruction model $g_\phi$ (e.g., an autoencoder) is trained to map each window $\mathbf{X}_{t}^{(W)}$ to a reconstruction $\widehat{\mathbf{X}}_{t}^{(W)}$:
\begin{equation}
\widehat{\mathbf{X}}_{t}^{(W)} = g_\phi(\mathbf{X}_{t}^{(W)}) \in \mathbb{R}^{W \times D}
\end{equation}

The reconstruction loss at time $t$ is defined as the Mean Squared Error (MSE) between the input and its reconstruction:
\begin{equation}
\label{eq:L_recon}
\mathcal{L}_{\text{recon}}(t) = \frac{1}{w \cdot D} \sum_{i=1}^{W} \sum_{j=1}^{D} \left(x_{t-W+i, j} - \widehat{x}_{t-W+i, j} \right)^2
\end{equation}

In our work, the raw anomaly scores, i.e., reconstruction error, predicted by the model are directly exploited for evaluation, overcoming the need for thresholding schemes to extract binary predictions. High reconstruction error values indicate points where the model fails to accurately reproduce the input, suggesting potential anomalies.

\subsection{Loss functions}
We experiment with different loss functions specifically designed for time series data. Considering the two anomaly detection approaches devised above, this leads to four separate configurations in our study.

\subsubsection{Mean Squared Error} This standard approach computes the loss based on the Mean Squared Error between the predicted sequence $\widehat{\mathbf{X}}$ obtained via the xLSTM decoder and the ground truth input sequence $\mathbf{X}$, as shown in Eq. \ref{eq:L_pred} \texttt{xLSTMAD-F} and\ Eq. \ref{eq:L_recon} \texttt{xLSTMAD-R}.

\subsubsection{Phase-Aware Shape Consistency Loss via SoftDTW}

To address the limitations of pointwise losses such as MSE in time series modeling, we incorporate a \textit{Soft Dynamic Time Warping (SoftDTW)} loss~\cite{cuturi2017soft}. Unlike traditional losses that penalize exact time-aligned discrepancies, SoftDTW enables differentiable sequence alignment, allowing for temporal shifts and warping, which are common in multivariate time series domains such as sensor networks, human activity recognition, and ECG signal monitoring.

Given an input sequence $\mathbf{X} \in \mathbb{R}^{T \times F}$ and its forecasted or reconstructed counterpart $\widehat{\mathbf{X}} \in \mathbb{R}^{T' \times F}$, we first define a cost matrix $C \in \mathbb{R}^{T \times T'}$ where each entry is the squared Euclidean distance between time step vectors:

\begin{equation}
C_{i,j} = \| \mathbf{X}_i - \widehat{\mathbf{X}}_j \|^2 \quad \text{for } i = 1,\dots,T; \; j = 1,\dots,T'
\end{equation}

The SoftDTW loss is then computed using a smooth approximation of the classical Dynamic Time Warping cost, defined recursively using the softmin operator:

\begin{equation}
\mathcal{D}_\gamma(i,j) = C_{i,j} + \gamma \cdot \text{softmin} \left(
\begin{array}{c}
\mathcal{D}_\gamma(i-1, j) \\
\mathcal{D}_\gamma(i, j-1) \\
\mathcal{D}_\gamma(i-1, j-1)
\end{array}
\right)
\end{equation}

where $\gamma > 0$ is a smoothing parameter, and the \textit{softmin} function is defined as:

\begin{equation}
    \begin{split}
\text{softmin}(a_1, a_2, a_3) & = 
-\gamma \cdot \log \exp\left( -\frac{a_1}{\gamma} \right) + \\ & + \exp\left( -\frac{a_2}{\gamma} \right) +  \exp\left( -\frac{a_3}{\gamma} \right) ) \\
\end{split}
\end{equation}

The final SoftDTW loss is the value of $\mathcal{D}_\gamma(T, T')$, which reflects the smoothed alignment cost between $\mathbf{X}$ and $\widehat{\mathbf{X}}$:

\begin{equation}
\mathcal{L}_{\text{SoftDTW}} = \mathcal{D}_\gamma(T, T')
\end{equation}

This formulation preserves differentiability with respect to both inputs and outputs, allowing seamless integration into gradient-based optimization. Moreover, by accounting for local temporal misalignments, SoftDTW provides a more robust metric for sequence similarity, especially in tasks such as anomaly detection, where phase variation may not correspond to actual anomalies.

\section{Experimental Setup}
In our experiments, we leverage the recently proposed comprehensive TSB-AD benchmark, which tackles the multiple issues affecting previous anomaly detection evaluations, such as flawed datasets, biased evaluation measures, and inconsistent benchmarking practices \cite{liu2024elephant}.
The benchmark includes a diverse set of multivariate time series datasets, with different underlying domain characteristics. 
This variety allows us to provide a comprehensive evaluation across various real-world challenges. In the following, we describe each dataset in more detail.

\subsection{Datasets}
The TSB-AD time series anomaly detection benchmark contains an extensive collection of 13 univariate and 20 multivariate public time-series anomaly detection datasets. In this study, we focus on the multivariate datasets:

\begin{itemize}
\item{\textbf{GHL} \cite{filonov2016multivariate}: Includes operational data from three reservoirs, capturing temperature and water level metrics. Anomalies are characterized by shifts in maximum temperature or deviations in pump frequency.}
\item{\textbf{Daphnet} \cite{bachlin2009wearable}: Comprises annotated accelerometer data from three sensors placed on Parkinson’s disease patients. The dataset focuses on detecting episodes of freezing of gait (FoG) during walking tasks.}
\item{\textbf{Exathlon} \cite{jacob2020exathlon}: Derived from real-world traces collected over 2.5 months from a Spark cluster. Each anomaly instance includes ground truth labels for both the root cause interval and the effect interval.}
\item{\textbf{Genesis} \cite{von2018anomaly}: A dataset from a portable pick-and-place demonstrator system that utilizes an air tank to operate gripping and storage components.}
\item{\textbf{Opportunity (OPP)} \cite{roggen2010collecting}: Consist of human activity data. It includes motion sensor data recorded as subjects performed typical daily activities.}
%
\item{\textbf{Server Machine Dataset (SMD)} \cite{su2019robust}: The dataset captures server metrics for anomaly detection, collected over a five-week period from 28 machines.}
\item{\textbf{SWaT} \cite{mathur2016swat}: A secure water treatment dataset with readings from 51 sensors and actuators. Anomalies in the dataset correspond to  attack scenarios.}
\item{\textbf{PSM} \cite{abdulaal2021practical}: Contains system-level metrics collected from various application server nodes at eBay.}
\item{\textbf{SMAP} \cite{hundman2018detecting}: Composed of telemetry data from NASA’s Soil Moisture Active Passive (SMAP) satellite.}
\item{\textbf{MSL}} \cite{hundman2018detecting}: Collected from the Curiosity Rover as part of Mars Science Laboratory (MSL) operations. The dataset includes telemetry data with annotated anomalies.
\item{\textbf{CreditCard} \cite{sharafaldin2018toward}: An intrusion detection dataset containing labeled network flows, including full packet payloads, traffic profiles, and flow labels.}
\item{\textbf{GECCO} \cite{moritz2018gecco}: A water quality dataset focused on the monitoring of drinking water safety.}
\item{\textbf{MIT-BIH Arrhythmia Database (MITDB)} \cite{goldberger2000physiobank}: Contains 48 half-hour two-channel ECG recordings from 47 individuals.}
\item{\textbf{Supraventricular Database (SVDB)} \cite{greenwald1990improved}:
Provides 78 half-hour ECG recordings that complement the MITDB by emphasizing supraventricular arrhythmias.}
\item{\textbf{Long-Term Database (LTDB)} \cite{goldberger2000physiobank}: A collection of seven extended-duration ECG recordings (14–22 hours each), featuring manually verified beat annotations.}
\item{\textbf{CATSv2} \cite{CATSV2}:  The second version of the Controlled Anomalies Time Series dataset, consisting of command sequences, external stimuli, and telemetry from a simulated complex system.}
\item{\textbf{TAO} \cite{TAO}: It contains 575,648 records with 3 attributes that are collected from the Tropical Atmosphere Ocean project.}
\end{itemize}

\subsection{Metrics}
Popular metrics in anomaly detection include threshold-based metrics (\textbf{Precision}, \textbf{Recall}, \textbf{F1-Score}) that allow the evaluation of the various aspects of the performance of the model. 
{ 
A commonly used extension of F1-Score is the \textbf{PA-F1} metric, which utilizes point adjustment on predictions. F1 has also been extended to consider the sequential nature of time series. The \textbf{event-based F1} that considers each anomaly segment as a distinct event, contributing exactly once as a true positive or a false negative. On the other hand, the\textbf{ R-based-F1} extends traditional evaluation metrics by incorporating components such as existence reward, overlap reward, and a cardinality factor. Moreover, \textbf{Affiliation-F1} emphasizes the temporal closeness between predicted and actual anomaly segments, quantifying their alignment based on the temporal distance \cite{liu2024elephant}.
}
However, the values of these metrics vary significantly depending on the choice of anomaly threshold. 

To evaluate a model using its corresponding anomaly scores, the Area Under the Receiver Operating Characteristics curve (\textbf{AUC-ROC}) \cite{fawcett2006introduction} is a much more desirable metric. It is defined as the area under the curve corresponding to TPR on the y-axis and FPR on the x-axis, as the threshold of the anomaly score varies. Another metric following a similar idea is the Area Under the Precision-Recall curve (\textbf{PR-AUC} \cite{davis2006relationship}), which leverages Precision on the y-axis and Recall on the x-axis.
Although these metrics overcome the limitation of threshold-based metrics, they only gather point-based anomalies.

Two desiderata in time series are \textit{range detection}, i.e., the anomaly detection algorithm should ideally detect every point in the anomalous sequence, and \textit{existence detection}, i.e., the detection of a tiny segment of one anomalous sequence is preferred rather than missing out on the whole sequence.
To address this problem, a promising direction was recently proposed in \cite{paparrizos2022volume}, where the AUC was extended for range-based anomaly detection. 

\textbf{Volume Under the Surface (VUS)} extends the mathematical model of Range-AUC measures by varying the window length, making the VUS family of measures truly parameter-free. 
This formulation also leads to an increased robustness to lag, noise, and anomaly cardinality ratio, as well as a high separability between accurate and inaccurate methods and consistency.

\textbf{VUS-ROC} and \textbf{VUS-PR} build a surface of TPR, FPR based on varying window lengths. The volume under the surface then represents a measure of AUC for various windows.
Since many time series anomaly detection models are sensitive to different window-length setups, the adoption of VUS enables a more fair and robust estimation and comparison of model performance.
Due to these powerful properties, we focus our experiments on the VUS family metrics while including the other classical metrics for completeness.  
A more formal definition of the VUS family of measures can be found in \cite{paparrizos2022volume}. 

\subsection{Reproducibility guidelines}
{

We follow the exact guidelines provided by the authors of the TSB-AD benchmark to ensure reproducibility and fair comparison of future methods. Although we provide a brief description of the evaluation scheme in the following, a detailed description and statistics can be found in the original paper \cite{liu2024elephant}. 
The hyperparameter tuning set consists of 15\% of each dataset, while the rest of the data is used to train and evaluate anomaly detection methods. The evaluation data is divided into 180 time series across 17 datasets. Each evaluation time series is then divided into training and test data sets. The datasets, along with information on all the splits, are available in the original TSB-AD repository: \url{https://github.com/TheDatumOrg/TSB-AD/}.

It should be noted that a single set of tuned hyperparameters is selected for all datasets rather than tuning hyperparameters specifically for each dataset. Such a decision improves the analysis of the models' robustness but makes it more challenging to achieve high results across all datasets. 

The hyperparameters search space for our methods includes:
\begin{itemize}
    \item \texttt{xLSTMAD-R}: window size: \{15, 25, 50\}, learning rate: \{$0.005$, $0.001$, $0.0008$\}, embedding dimensionality: \{20, 40\}.
    \item \texttt{xLSTMAD-F}: window size: \{15, 25, 50\}, learning rate: \{$0.005$, $0.001$, $0.0008$\}, embedding dimensionality: \{20, 40\}, forecasting horizon: \{1, 5, 15\}.
\end{itemize}

We selected the hyperparameters that yield the highest value of the VUS-PR metric achieved on the tuning set. The final hyperparameters are presented in Table \ref{tab:hyperparameters}.

\begin{table}[]
    \centering
    \caption{Final hyperparameters for all methods: Window Size (Win. Size), Learning Rate (LR), Embedding Dimensionality (Embed. Dim), Forecasting Horizon (Horizon). NA = Non Applicable}
    \label{tab:hyperparameters}
    \begin{tabular}{cccccc}
    \toprule
         Method & Win. Size & LR & Embed. Dim & Horizon \\
         \midrule
         xLSTM-F (MSE) & 50 & 0.0008 & 20 & 5 & \\
        xLSTM-F (DTW) & 25 & 0.005 & 20 & 5 & \\
        xLSTM-R (MSE) & 50 & 0.005 & 40 & NA & \\
        xLSTM-R (DTW) & 25 & 0.001 & 20 & NA & \\

    \bottomrule
    \end{tabular}
\end{table}

The evaluation set is divided into 180 time series that are then split into training and test parts. As a result, our experiments included 720 training and test executions for our \texttt{xLSTMAD-R} and \texttt{xLSTMAD-F} methods with two losses (MSE and SoftDTW). All experiments were performed on a single Nvidia GH200 GPU.  Our code is available at \url{https://github.com/Nyderx/xlstmad}.

}






\section{Results Discussion}
Our experiments are aimed at answering the following research questions:
\begin{itemize}
\item{\textbf{RQ1.} Do our proposed \texttt{xLSTMAD} strategies exhibit superior performance on the challenging TSB-AD-M benchmark compared to state-of-the-art baselines in anomaly detection?} 
\item{\textbf{RQ2.} Does \texttt{xLSTMAD} consistently outperform other methods in a fine-grained comparison that considers single datasets?}
\item{\textbf{RQ3.} Are there sensible differences in predictive behavior between the proposed anomaly detection losses (MSE vs. SoftDTW) and anomaly detection approaches (\texttt{xLSTMAD-F} vs. \texttt{xLSTMAD-R})?}
\end{itemize}

\begin{table*}[!ht]
\setlength{\tabcolsep}{4pt}
    \caption{Summary comparison of our proposed method with 23 competitors across 180 time-series originating from 17 datasets in terms of multiple metrics. The best-performing method for each metric is marked in bold.}
    \centering
    \begin{tabular}{llllllllll}
    \toprule
Method & VUS-PR & VUS-ROC & AUC-PR & AUC-ROC & Standard-F1 & PA-F1 & Event-based-F1 & R-based-F1 & Affiliation-F1\\ 
\midrule
xLSTMAD-F (MSE) & 0.35 & \textbf{0.77} & \textbf{0.35} & \textbf{0.74} & \textbf{0.40 }& \textbf{0.85 }& \textbf{0.70} &\textbf{ 0.42} & \textbf{0.89}\\ 
xLSTMAD-F (SoftDTW) & 0.34 & 0.76 & \textbf{0.35} & 0.73 & \textbf{0.40} & 0.83 & 0.67 & 0.41 & 0.88  \\ 
xLSTMAD-R (MSE) & \textbf{0.37} & 0.72 & 0.32 & 0.68 & 0.38 & 0.53 & 0.45 & 0.36 & 0.82 \\ 
xLSTMAD-R (SoftDTW) & 0.36 & 0.72 & 0.31 & 0.68 & 0.37 & 0.57 & 0.48 & 0.35 & 0.82 \\ 
RandomModel & 0.10 & 0.59 & 0.05 & 0.50 & 0.09 & 0.71 & 0.10 & 0.10 & 0.69 \\ 
\midrule
CNN \cite{munir2018deepant}& 0.31 & 0.76 & 0.32 & 0.73 & 0.37 & 0.78 & 0.65 & 0.37 & 0.87 \\ 
OmniAnomaly \cite{su2019robust} & 0.31 & 0.69 & 0.27 & 0.65 & 0.32 & 0.55 & 0.41 & 0.37 & 0.81 \\ 
PCA & 0.31 & 0.74 & 0.31 & 0.7 & 0.37 & 0.79 & 0.59 & 0.29 & 0.85 \\ 
LSTMAD \cite{ji2021novel} & 0.31 & 0.74 & 0.31 & 0.7 & 0.36 & 0.79 & 0.64 & 0.38 & 0.87 \\ 
USAD \cite{audibert2020usad} & 0.3 & 0.68 & 0.26 & 0.64 & 0.31 & 0.53 & 0.4 & 0.37 & 0.8 \\ 
AutoEncoder \cite{sakurada2014anomaly} & 0.3 & 0.69 & 0.3 & 0.67 & 0.34 & 0.6 & 0.44 & 0.28 & 0.8 \\ 
KMeansAD \cite{yairi2001fault} & 0.29 & 0.73 & 0.25 & 0.69 & 0.31 & 0.68 & 0.49 & 0.33 & 0.82 \\ 
CBLOF \cite{he2003discovering} & 0.27 & 0.7 & 0.28 & 0.67 & 0.32 & 0.65 & 0.45 & 0.31 & 0.81 \\ 
MCD \cite{estimator1999fast} & 0.27 & 0.69 & 0.27 & 0.65 & 0.33 & 0.46 & 0.33 & 0.2 & 0.76 \\ 
OCSVM \cite{hejazi2013one} & 0.26 & 0.67 & 0.23 & 0.61 & 0.28 & 0.48 & 0.41 & 0.3 & 0.8 \\ 
Donut \cite{xu2018unsupervised} & 0.26 & 0.71 & 0.2 & 0.64 & 0.28 & 0.52 & 0.36 & 0.21 & 0.81 \\ 
RobustPCA \cite{paffenroth2018robust} & 0.24 & 0.61 & 0.24 & 0.58 & 0.29 & 0.6 & 0.42 & 0.33 & 0.81 \\ 
FITS \cite{xufits} & 0.21 & 0.66 & 0.15 & 0.58 & 0.22 & 0.72 & 0.32 & 0.16 & 0.81 \\ 
OFA \cite{zhou2023one} & 0.21 & 0.63 & 0.15 & 0.55 & 0.21 & 0.72 & 0.41 & 0.17 & 0.83 \\ 
EIF \cite{hariri2019extended} & 0.21 & 0.71 & 0.19 & 0.67 & 0.26 & 0.74 & 0.44 & 0.26 & 0.81 \\ 
COPOD \cite{li2020copod} & 0.2 & 0.69 & 0.2 & 0.65 & 0.27 & 0.72 & 0.41 & 0.24 & 0.8 \\ 
IForest \cite{liu2008isolation} & 0.2 & 0.69 & 0.19 & 0.66 & 0.26 & 0.68 & 0.41 & 0.24 & 0.8 \\ 
HBOS \cite{goldstein2012histogram}& 0.19 & 0.67 & 0.16 & 0.63 & 0.24 & 0.67 & 0.4 & 0.24 & 0.8 \\ 
TimesNet \cite{wutimesnet} & 0.19 & 0.64 & 0.13 & 0.56 & 0.2 & 0.68 & 0.32 & 0.17 & 0.82 \\ 
KNN \cite{ramaswamy2000efficient} & 0.18 & 0.59 & 0.14 & 0.51 & 0.19 & 0.69 & 0.45 & 0.21 & 0.79 \\ 
TranAD \cite{tuli2022tranad} & 0.18 & 0.65 & 0.14 & 0.59 & 0.21 & 0.68 & 0.4 & 0.21 & 0.79 \\ 
LOF \cite{breunig2000lof} & 0.14 & 0.6 & 0.1 & 0.53 & 0.15 & 0.57 & 0.32 & 0.14 & 0.76 \\ 
AnomalyTransformer \cite{xuanomaly} & 0.12 & 0.57 & 0.07 & 0.52 & 0.12 & 0.53 & 0.33 & 0.14 & 0.74\\

        \bottomrule
    \end{tabular}
\label{tbl:results-general}
\end{table*}



\begin{table*}[!ht]
\setlength{\tabcolsep}{4pt}
\caption{Comparison of our proposed method with its closer baseline, LSTMAD, the best other performing model, CNN, and a model returning random predictions (Rand). We present the results for the two most comprehensive metrics, VUS-PR and VUS-AUC. Our method is presented in two variants: reconstruction (noted as \textbf{R}) and forecasting (noted as \textbf{F}) with two losses: \textbf{MSE} and SoftDTW (noted as \textbf{DTW}).The best-performing method for each metric and dataset is marked in bold.}
    \centering
    \begin{tabular}{lll|ll|ll|l||ll|ll|ll|ll}
    \toprule
& \multicolumn{7} {c||} { VUS-PR } &  \multicolumn{7} {c} { VUS-ROC} \\
\midrule
 Dataset & R(MSE) & R(DTW) & F(MSE) & F(DTW) & LSTMAD & CNN & Rand & R(MSE) & R(DTW) & F(MSE) & F(DTW) & LSTM & CNN & Rand \\
\midrule 
CATSv2 \cite{CATSV2} & \textbf{0.10}& \textbf{0.10} & 0.04 & 0.04 & 0.04 & 0.08 & 0.03 & 0.61 & 0.61 & 0.46 & 0.46 & 0.43 & \textbf{0.64} & 0.53 \\
CreditCard \cite{sharafaldin2018toward} & 0.06 &\textbf{0.07} & 0.02 & 0.02 & 0.02 & 0.02 & 0.02 & 0.75 & 0.81 & \textbf{0.87} &\textbf{0.87} & \textbf{0.87} & 0.86 & 0.84 \\
Daphnet \cite{bachlin2009wearable} & \textbf{0.50} & 0.46 & 0.31 & 0.31 & 0.31 & 0.21 & 0.06 & \textbf{0.95 }& \textbf{0.95} & 0.81 & 0.81 & 0.81 & 0.84 & 0.52 \\
Exathlon \cite{jacob2020exathlon} & \textbf{0.91} &\textbf{0.91} & 0.82 & 0.82 & 0.82 & 0.68 & 0.10 & \textbf{0.95} &\textbf{0.95} & 0.89 & 0.89 & 0.89 & 0.88 & 0.51 \\
GECCO \cite{moritz2018gecco} & \textbf{0.04 }&\textbf{0.04} & 0.02 & 0.02 & 0.02 & 0.03 & 0.02 & 0.50 & 0.54 & 0.44 & 0.44 & 0.44 &\textbf{0.65} & 0.64 \\
GHL \cite{filonov2016multivariate} & 0.01 & 0.01 & \textbf{0.16} & \textbf{0.16} & 0.06 & 0.02 & 0.01 & 0.46 & 0.45 & \textbf{0.69} & 0.68 & 0.63 & 0.55 & 0.52 \\
Genesis \cite{von2018anomaly} & 0.01 & 0.01 & 0.03 & 0.03 & 0.04 & \textbf{0.10} & 0.01 & 0.78 & 0.85 & 0.58 & 0.58 & 0.58 & \textbf{0.96} & 0.77 \\
LTDB \cite{goldberger2000physiobank} & \textbf{0.40} & 0.34 & 0.34 & 0.34 & 0.31 & 0.33 & 0.19 & 0.70 & 0.68 & 0.71 & 0.71 & 0.71 & \textbf{0.72} & 0.58 \\
MITDB \cite{goldberger2000physiobank} & 0.09 & 0.08 & 0.13 & 0.13 & 0.11 & \textbf{0.14} & 0.04 & 0.66 & 0.66 & \textbf{0.69} & 0.68 & 0.67 &\textbf{0.69} & 0.54 \\
MSL \cite{hundman2018detecting} & 0.38 & 0.40 & \textbf{0.41} & 0.34 & 0.23 & 0.35 & 0.08 & \textbf{0.81 }& \textbf{0.81 }& \textbf{0.81} & 0.77 & 0.73 & 0.77 & 0.63 \\
OPPORT. \cite{roggen2010collecting} & 0.16 & 0.16 & 0.16 & 0.16 & \textbf{0.17} & 0.16 & 0.05 & 0.28 & 0.28 & \textbf{0.65} &\textbf{0.65} &\textbf{0.65} & 0.61 & 0.53 \\
PSM \cite{abdulaal2021practical} & 0.18 & 0.18 & \textbf{0.24} &\textbf{0.24} & 0.22 & 0.22 & 0.13 & 0.61 & 0.62 & \textbf{0.74 }& 0.73 & 0.72 & 0.70 & 0.54 \\
SMAP \cite{hundman2018detecting} & \textbf{0.35} & 0.32 & 0.20 & 0.22 & 0.17 & 0.20 & 0.04 & 0.73 & 0.74 & 0.76 & 0.76 & 0.71 &\textbf{0.78} & 0.60 \\
SMD \cite{su2019robust} & 0.35 & \textbf{0.37} & 0.31 & 0.30 & 0.31 & 0.37 & 0.05 & 0.83 & \textbf{0.84} & 0.83 & 0.83 & 0.83 & 0.83 & 0.60 \\
SVDB \cite{greenwald1990improved} &\textbf{0.26} & 0.19 & 0.20 & 0.20 & 0.15 & 0.19 & 0.06 & 0.68 & 0.67 & 0.72 & 0.72 & 0.69 &\textbf{0.73} & 0.57 \\
SWaT \cite{mathur2016swat} & 0.34 & \textbf{0.35 }& 0.16 & 0.16 & 0.16 & 0.48 & 0.14 & 0.71 & 0.71 & 0.50 & 0.50 & 0.47 & \textbf{0.74} & 0.53 \\
TAO \cite{TAO} & 0.80 & 0.82 & \textbf{1.00} & \textbf{1.00} & 0.99 & \textbf{1.00} & 0.77 & 0.88 & 0.91 & \textbf{1.00} &\textbf{1.00} & \textbf{1.00} & \textbf{1.00} & 0.94 \\
\bottomrule
    \end{tabular}
\label{tbl:results-fine-grained}
\end{table*}

{

We recall that challenging experimental choices originating from the adopted benchmark may impact the performance of each model. Specifically, hyperparameter tuning is performed just once, selecting a single hyperparameter configuration that achieves the best results across all datasets. 
%
Another observation is that VUS-PR and VUS-ROC are very challenging metrics. Since they evaluate model performance across multiple time periods, they exacerbates the evaluation for all models, yielding significantly lower results than ROC-AUC. 
%

In the following discussions, we focus on comparisons between our proposed method and two baselines: CNN \cite{munir2018deepant} and LSTMAD \cite{ji2021novel}. While CNN is relevant since it achieved the best performance in the TSB-AD benchmark \cite{liu2024elephant}, LSTMAD is a natural competitor since it is based on LSTM, the predecessor of xLSTM. 
Moreover, to support the interpretation of experimental results, we provide a Random model that simply returns a random anomaly score in the range from 0 to 1. This allows us to set realistic expectations for VUS-based metric values for non-random anomaly detection methods.

\subsection{xLSTMAD results across all datasets}
Table~\ref{tbl:results-general} shows that all \texttt{xLSTMAD} variants outperform state-of-the-art baselines, especially in terms of the most challenging VUS-PR metric.
%
The best performing \texttt{xLSTMAD-R} with MSE loss outperforms current state-of-the-art results, achieved by CNN, OmniAnomaly, and LSTMAD, by almost 20\% (0.37 vs 0.31). In particular, it performs 370\% better than the random model (0.37 vs. 0.10).
%

While \texttt{xLSTMAD-R} presents better results than other variants in terms of VUS-PR (0.37), the proposed \texttt{xLSTMAD-F} model with MSE loss consistently outperforms a broad range of classical and deep learning baselines across all metrics. It achieves the highest AUC-PR (0.35), AUC-ROC (0.74), VUS-ROC (0.77), Affiliation-F1 (0.89), PA-F1 (0.85), Event-based-F1 (0.70), R-Based-F1 (0.42), and Standard-F1 (0.40), indicating robust performance across both classical detection (AUC-PR/ROC) and other anomaly detection metrics.  

Although CNN and LSTMAD are competitive in some individual metrics (e.g., CNN reaches Event-based-F1 of 0.65), they are consistently lower than \texttt{xLSTMAD-F} in overall anomaly structure modeling. Notably, classical methods such as PCA, IForest, or AutoEncoder perform significantly worse across all detection metrics. 

These results highlight the strength of \texttt{xLSTMAD} on this challenging multivariate anomaly detection benchmark \textbf{(RQ1)}.




\subsection{Fine-grained anomaly detection results} 
%
%
%
%
Table \ref{tbl:results-fine-grained} shows the fine-grained results for each of the 17 datasets from the TSB-AD-M benchmark, where our proposed \texttt{xLSTMAD} method is compared in its four variants (\texttt{xLSTMAD-R MSE}, \texttt{xLSTMAD-R SoftDTW}, \texttt{xLSTM-F MSE}, \texttt{xLSTMAD-F SoftDTW}}) against the top-performing methods in the general results across multiple datasets (LSTM, CNN).
This allows us to achieve a fine-grained view of model performance on specific datasets, discarding low-performing baselines.
In addition to anomaly detection baselines, we display performance obtained with the Random model (Rand) as a reference point for random performance. Its VUS-PR scores are close to zero for most datasets, with the exception of TAO.

%
%
%
The results show that our models consistently outperform the LSTM and CNN baselines across most data sets in both VUS-PR and VUS-ROC. Notably, \texttt{xLSTMAD-R} often dominates in VUS-PR, while \texttt{xLSTMAD-F} shines in VUS-ROC, suggesting complementary strengths between reconstruction and prediction-based detection.
On \textit{industrial datasets} such as Exathlon, SMD, and SWaT, our models achieve strong performance, with \texttt{xLSTMAD-R} matching or exceeding baselines for both MSE and SoftDTW loss. For Exathlon, \texttt{xLSTMAD-R} (SoftDTW) reaches a near-perfect VUS-PR of 0.91 and VUS-ROC of 0.95.
In \textit{physiological datasets} like Daphnet, MITDB, and LTDB, our models show notable improvements. 

On Daphnet, \texttt{xLSTMAD-R} (MSE) achieves a VUS-PR of 0.50 compared to 0.31 for LSTM and 0.21 for CNN.
On \textit{high-noise or sparse} anomaly datasets such as CATSv2 and GECCO, the performance margins are smaller, though \texttt{xLSTMAD-R} (SoftDTW) still outperforms CNN and LSTM in most cases. On the other hand, all variants of \texttt{xLSTMAD} achieve very low VUS-PR results for the Genesis dataset. The reason behind such results may be the fact that the Genesis dataset consists of just three very short anomalies \cite{liu2024elephant}.
The \textit{TAO dataset} is a notable example with perfect or near-perfect scores across all models. All four variants of our approach achieve VUS-ROC = 1.00 and two of them achieve VUS-PR = 1.00 \texttt{xLSTMAD-F}, indicating the ease of detecting anomalies in this scenario.

In summary, the proposed xLSTM-based methods consistently demonstrate high detection capability across various domains, delivering the best results in terms of VUS-PR in 14 of 17 datasets (\textbf{RQ2}).

\subsection{Comparison of \texttt{xLSTMAD} variants}


A closer inspection of the four \texttt{xLSTMAD} variants reveals distinct behavioral patterns.

\textbf{Forecasting vs. Reconstruction modes:} Based on results in Table \ref{tbl:results-general}, the reconstruction-based variant \texttt{xLSTMAD-R} leads in average performance according to VUS-PR. However, the forecasting-based variant \texttt{xLSTMAD-F} outperforms \texttt{xLSTMAD-R} across nearly all other metrics, showing complementary strengths, especially when temporal alignment is critical. For example, \texttt{xLSTMAD-F} (MSE) achieves 0.70 Event-based-F1 vs. 0.45 for \texttt{xLSTMAD-R} (MSE), and 0.89 Affiliation-F1 vs. 0.82, suggesting that forecasting-based anomaly detection is more effective at modeling latent temporal expectations than reconstruction-based approaches.

\textbf{Loss Type (MSE vs. SoftDTW):} The SoftDTW-based loss offers competitive gains, particularly in reconstruction mode in terms of VUS-ROC, for some datasets such as CreditCard (from 0.75 to 0.81), GECCO (from 0.50 to 0.54), and TAO (from 0.88 to 0.91), due to their ability to accommodate small temporal shifts and warping in sequences.
However, it can be observed that the use of SoftDTW does not consistently outperform MSE. From a more global perspective, for both forecasting and reconstruction modes, the MSE variant exhibits marginally better results across most metrics, such as AUC-PR and Event-based-F1. 
This indicates that while SoftDTW introduces alignment-aware sensitivity, it may also increase noise sensitivity or reduce stability during training.

In summary, these comparisons emphasize that both model design and loss function selection play crucial roles in multivariate anomaly detection performance, with \texttt{xLSTMAD-F} and MSE emerging as the most robust combination on the TSB-AD-M benchmark (\textbf{RQ3}).

\subsection{Empirical time complexity}
We analyze the average execution time (in seconds, using Nvidia GH200 GPU) of different anomaly detection methods on all 180 time series originating from 17 datasets (including training and testing steps). As expected, reconstruction-based approaches incur the highest computational cost of 626 and 355 seconds on average for MSE and SoftDTW losses, respectively. 
This computational cost stems mostly from the complexity of reconstructing multivariate sequences (window size of 50 and 25 time steps, respectively, see Table \ref{tab:hyperparameters}).
%
Prediction-based models offer better efficiency (121 seconds for MSE loss and 264 seconds for SoftDTW loss). This improvement is related to the fact that instead of reconstructing the whole window, the model predicts only time steps from a much shorter forecasting horizon (5 time steps in our experiments).

In contrast, baseline deep learning models like LSTM and CNN are significantly faster, with CNN achieving the lowest execution time of just 17.12 seconds. This highlights a clear trade-off between detection performance and computational efficiency: while xLSTM-based methods (especially those using DTW) provide robust anomaly detection performance, they do so at a higher computational cost. However, it should be noted that xLSTM models are still in their infancy, and improvements in their efficiency may be achieved with future implementations that simplify the architecture or leverage hardware acceleration more extensively. 




\section{Conclusion}
In this paper, we propose \texttt{xLSTMAD}, the first method that integrates a full encoder-decoder xLSTM architecture, purpose-built for multivariate time series anomaly detection.
%
Following the two most adopted types of approach to anomaly detection, our method is designed in two variants: forecasting \texttt{xLSTMAD-F} and reconstruction \texttt{xLSTMAD-R}, which provide a high degree of flexibility to different domains.
Furthermore, we investigate the performance of two loss functions: Mean Squared Error (MSE), and Soft Dynamic Time Warping (SoftDTW) to consider local reconstruction fidelity and global sequence alignment, respectively. 
In our results on the comprehensive TSB-AD-M benchmark, 
xLSTM outperforms 23 popular anomaly detection baselines, achieving state-of-the-art performance in terms of challenging metrics such as VUS-PR and VUS-ROC. 
%
%

As future work, we aim to further study xLSTM modeling capabilities in the context of anomaly detection, leading to new model architectures that are capable of providing a more balanced performance-efficiency tradeoff.

\section*{Acknowledgments}
Project cofunded by: i) the Polish Ministry of Science and Higher Education under the program "Co-funded International Projects,"; ii) the program "Excellence initiative-research university" for the AGH University in Krakow; iii) the ARTIQ project: UMO-2021/01/2/ST6/00004 and ARTIQ/0004/2021.

\bibliographystyle{IEEEtran}
\bibliography{refs}

\end{document}